\documentclass[runningheads]{llncs}

\usepackage{eccv}

\usepackage{eccvabbrv}

\usepackage{graphicx}
\usepackage{booktabs}

\usepackage[accsupp]{axessibility}  

\usepackage{multirow}

\usepackage{hyperref}

\usepackage{orcidlink}
\usepackage{graphicx}
\usepackage{bbm}
\usepackage[symbol]{footmisc}

\begin{document}

\title{OAT: Object-Level Attention Transformer for Gaze Scanpath Prediction} 

\titlerunning{Object-level Attention Transformer}

\author{Yini Fang\inst{*1}\orcidlink{0009-0008-9478-5545} \and
Jingling Yu\inst{*1}\orcidlink{0000-0002-3076-1287} \and
Haozheng Zhang\inst{2}\orcidlink{0000-0003-1312-4566} \and
Ralf van der Lans\inst{1}\orcidlink{0000-0002-7726-8238} \and
Bertram Shi \inst{1}\orcidlink{0000-0001-9167-7495}
}

\authorrunning{Y. Fang et al.}

\institute{Hong Kong University of Science and Technology, Clear Water Bay, HK  \and
University of Durham, Durham, UK
\\
\email{\{yfangba, jyubj\}@connect.ust.hk}, \email{haozheng.zhang@durham.ac.uk}, \email{rlans@ust.hk}, \email{eebert@ust.hk}
}

\maketitle

\begin{abstract}
Visual search is important in our daily life. The efficient allocation of visual attention is critical to effectively complete visual search tasks. Prior research has predominantly modelled the spatial allocation of visual attention in images at the pixel level, e.g. using a saliency map. However, emerging evidence shows that visual attention is guided by objects rather than pixel intensities. This paper introduces the Object-level Attention Transformer (OAT), which predicts human scanpaths as they search for a target object within a cluttered scene of distractors. OAT uses an encoder-decoder architecture. The encoder captures information about the position and appearance of the objects within an image and about the target. The decoder predicts the gaze scanpath as a sequence of object fixations, by integrating output features from both the encoder and decoder. We also propose a new positional encoding that better reflects spatial relationships between objects. We evaluated OAT on the Amazon book cover dataset and a new dataset for visual search that we collected. OAT's predicted gaze scanpaths align more closely with human gaze patterns, compared to predictions by algorithms based on spatial attention on both established metrics and a novel behavioural-based metric. Our results demonstrate the generalization ability of OAT, as it accurately predicts human scanpaths for unseen layouts and target objects. The code is available at: \url{https://github.com/HKUST-NISL/oat_eccv24}.

\footnotetext[1]{The first two authors contributed equally in the paper.}


\end{abstract}

\section{Introduction}
\label{sec:intro}




\begin{figure*}[t]
\includegraphics[width=0.8\textwidth]{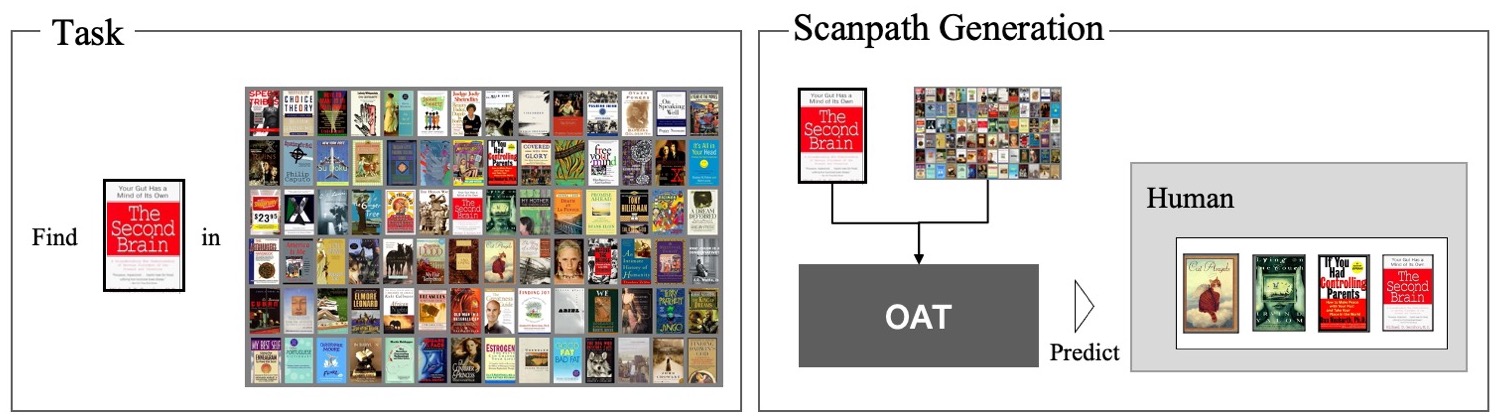}
\centering
\caption{Illustration of object-level scanpath prediction.}
\label{task}
\end{figure*}

Target search happens regularly in daily life. For example, parents may search for their children in the crowd at an amusement park. Short-sighted people may look for their glasses in the clutter of objects on their desk. Shoppers may search for a desired product from among an array of similar products on a shelf. Predicting people's gaze behaviour in visual search is vital, as it offers valuable implications across multiple domains, such as estimating user intentions and subsequently facilitating robot guidance in hierarchical tasks in HCI \cite{shen2023intention}, enhancing user experience in virtual reality (VR) and augmented reality (AR) environments \cite{bennett2021assessing, krajancich2020optimizing, rivu2020stare}, and increasing revenue through advertising and marketing strategies \cite{van2021online, wedel2019eye}.



Because the high-resolution vision that facilitates target identification is restricted to a small area (the fovea) of the eye, target search is often executed as a sequence of saccades to different locations in the visual field. This sequence of locations is referred to as the scanpath. Better understanding of visual search can be obtained through models that predict human scanpaths during visual search. 

Scanpath prediction is closely related to visual attention, a set of processes that identifies salient (important or task-relevant) areas in the visual field. Early models of scanpath prediction during free-viewing of images were based on the concept of two-dimensional saliency map, which assigns to each pixel a saliency value. Most work has focused on modelling free-viewing, where scanpaths are often generated by successively selecting pixels with high salience with an inhibition-of-return mechanism to avoid the same point being selected over and over again. Recently, models have begun to address scanpath prediction in task-oriented or goal-directed behaviors, like visual search \cite{zelinsky2019benchmarking, yang2020predicting, chen2021predicting,  mondal2023gazeformer, yang2023predicting}. These usually employ a spatial component for extracting visual features of the stimuli and a temporal component for generating the gaze sequence, but have retained a focus on pixel-level prediction.

However, increasing evidence suggests visual attention is not directed towards pixels, but rather objects, especially in the case of goal-directed behavior \cite{einhauser2008objects, nuthmann2010object, pajak2013object}. A wide range of psychophysical experiments have demonstrated the importance of objects in understanding visual attention \cite{kahneman1992reviewing, egly1994shifting, o1999fmri}. 

We hypothesize that modelling scanpaths as sequences of object fixations, rather than pixel fixations, will result in more accurate models of human gaze behavior, especially in tasks such as visual search. We anticipate that this approach will be especially valuable in handling cluttered environments containing many similar objects. As a step in this direction, we propose the \textbf{O}bject \textbf{A}ttention Modelling \textbf{T}ransformer \textbf{OAT} architecture for scanpath prediction at the object level. It  
consists of four modules: object embedding, encoder, decoder, and object attention. The object embedding module extracts visual and geometrical features for each object. For visual search, the transformer encoder combines information about the target and the objects and their positions in the scene. The transformer decoder assigns to each object the probability that it will be the next fixated object, based on cross attention computed between the past sequence of fixated objects and the encoder representation of the target and scene. Predictions of the next fixation can be produced by sampling from this distribution. Figure \ref{task} illustrates our framework using an example from the Amazon book cover dataset \cite{Stauden18}, where users search for a target book cover in a 2D array of book covers.

Our contributions can be summarized as follows:
\begin{enumerate}
  \item To the best of our knowledge, we propose the first approach to predict gaze scanpaths at the object level. We study visual search, but believe that this architecture is a valuable starting point for object-level modelling of other tasks.
  \item We propose two modifications to the transformer architecture to optimize performance on this task. The first is a 2D positional encoding algorithm that better captures the spatial relationships between objects. The second is a scalable object attention module that enables OAT to generate predictions for scenes containing varying numbers of objects without retraining. 
  \item For model validation, we propose a new metric for comparing generated scanpaths with human scanpaths, which compares the distributions of task-relevant gaze behaviors based on psychological processes of visual attention. 
  \item OAT achieves state-of-the-art performance compared to baselines using both previously proposed standard metrics, as well as our new metric. Our predicted scanpaths differ by only 3.1\% from human scanpaths.
\end{enumerate}

\section{Related Work}

While there is no work on gaze scanpath prediction at object level, there has been extensive investigation into gaze scanpath prediction at the pixel level. Early work by \cite{itti1998model} on the visual attention system and the development of saliency benchmarks \cite{judd2009learning, judd2012benchmark, kummerer2018saliency} have paved the way for this research. Most of the existing work in gaze scanpath modelling \cite{schwinn2022behind,linardos2021deepgaze,lou2022transal,cornia2018predicting} focuses on predicting free-viewing behaviour, where human attention is driven solely by visual input without a specific target. These models generate saliency maps to capture spatial dependencies, but do not adequately consider temporal dependencies in scanpaths.

In goal-directed scanpath prediction, attention is guided by a top-down goal, and sequence modelling algorithms are used to generate the sequence while capturing correlations between the target and the sequence. Zelinsky et al. \cite{zelinsky2019benchmarking, zelinsky2021predicting} employed Inverse Reinforcement Learning (IRL) to learn a target-specific reward function and evaluated their model on a self-collected dataset based on the COCO dataset. Yang et al. \cite{yang2020predicting} extended the IRL algorithm and tested it on the larger-scale COCO-Search18 dataset. They later introduced a more generalized scanpath prediction model that incorporates a transformer-based architecture and a simulated foveated retina to emulate human dynamic visual working memory \cite{yang2023predicting}. Mondal et al. \cite{mondal2023gazeformer} also applied a transformer-based architecture for goal-directed scanpath prediction, combining visual and semantic features to decode fixations. Tononi et al. \cite{tonini2023object} propose a Transformer-based architecture for gaze target detection, but focusing on predicting a single point a person inside the image is looking at, rather than predicting the scanpath of an external observer.


Previous models of goal-directed search have primarily focused on predicting scanpaths at the pixel level, rather than at the object level, as we propose. Pixel-based models may suffice for natural-scene images where objects are distinct and viewers can rely on prior knowledge about likely spatial locations. However, in more challenging scenarios such as scenes with random clutter, disorganization, or similar-looking distractors, pixel-based models may not be as appropriate. These difficult scenarios, as exemplified by the Amazon book cover dataset, 
require more fixations (8.95) on average compared to the COCO-Search18 dataset (3.5 fixations). The experiments conducted in this study demonstrate that object-level modeling outperforms pixel-level modeling in these challenging scenarios.

Our model is based on a transformer architecture similar to that proposed by Mondal et al. \cite{mondal2023gazeformer}. However, OAT introduces distinct differences. Firstly, it incorporates a target-dependent encoding of the input image by including the target as an extra token in the encoder transformer. This allows objects similar to the target to be encoded differently from dissimilar objects. Secondly, OAT includes past fixations in the decoder input instead of predicting all fixations at once. Lastly, OAT's object attention module predicts the fixated object, rather than the fixated pixel. 

\section{OAT Architecture}

\begin{figure*}[t]
 \includegraphics[width=0.88\textwidth]{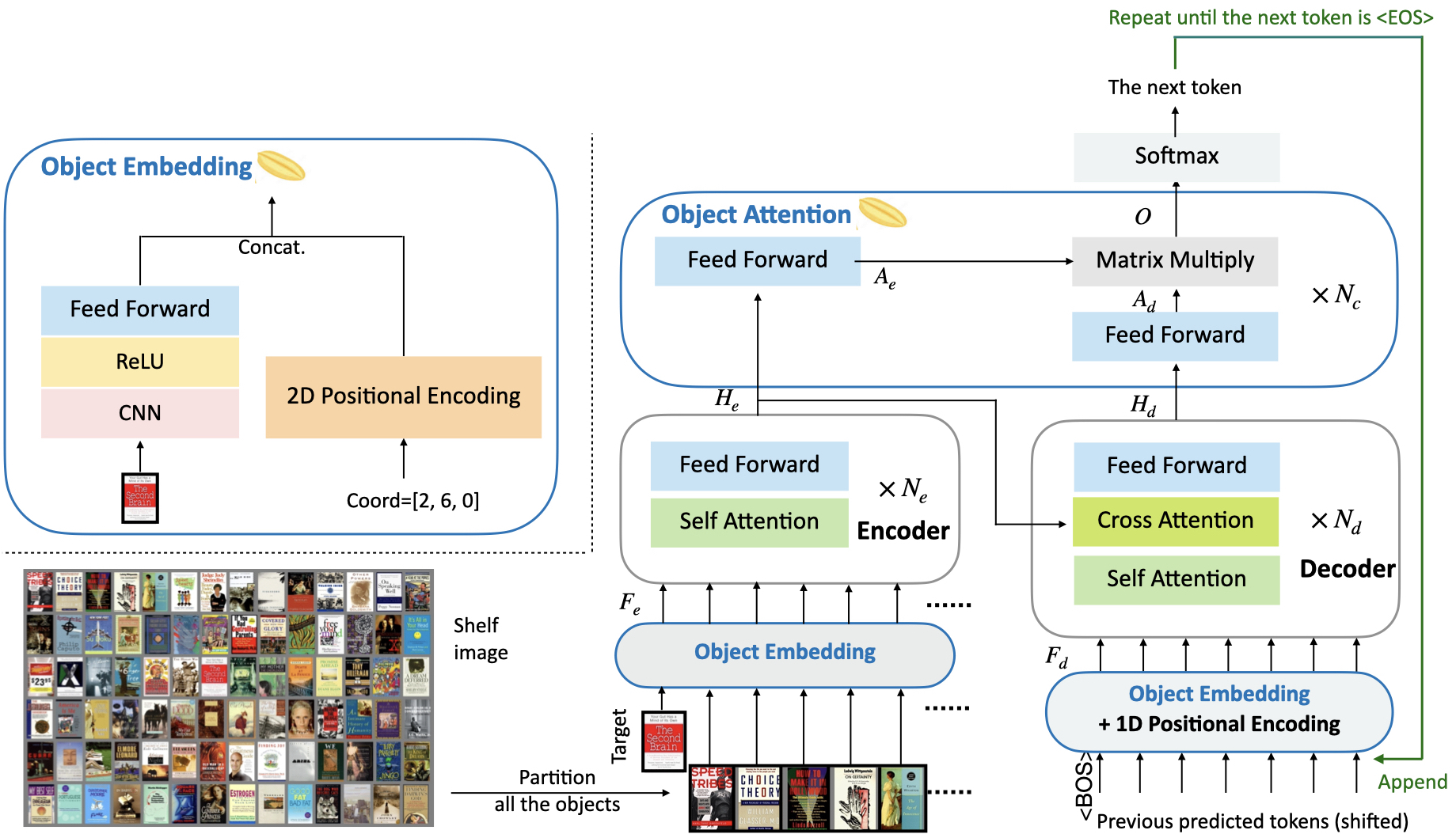}
\centering
\caption{An overview of OAT. The input to the encoder is a target object, an image containing the target object and other distractor objects, and the input to the decoder is the sequence of previously and currently fixated objects. The output is a probability distribution over objects being fixated next. The model repeats the process until the next token is the end token <EOS>. This is the process of predicting one gaze scanpath.}
\label{model1}
\end{figure*} 

Figure \ref{model1} presents an overview of OAT. It consists of 
an object embedding module, 
$N_e$ transformer encoder blocks and 
$N_d$ decoder blocks, and $N_c$ object attention blocks. The encoder and decoder architecture follow \cite{vaswani2017attention}. We first partition the image into objects and append the target object to the beginning of the object sequence. The object embedding module generates the embedding sequence $F_e$, which is then processed by encoder blocks to obtain the encoder hidden states $H_e$. The decoder computes the cross attention between $H_e$ and the embedding sequence from the decoder $F_d$, and outputs the hidden states of the decoder $H_d$. $H_e$ and $H_d$ are processed by the object attention module, which yields an output $O$. Finally, this is converted to a probability distribution over the next fixated object by the Softmax function.

\paragraph{Object Embedding and Encoder.}

The object embedding module computes a $p$-dimensional visual descriptor for each object. For the databases we consider here, objects are placed in regular arrays. Each object corresponds to a rectangular patch. We use a Convolutional Neural Network (CNN) followed by ReLU and feed-forward layers applied to the patch to extract visual features. More complex layouts, where objects have variable sizes and potentially overlap/occlude, can be handled by first segmenting the image, then extracting object visual descriptors. 

For each object, we also compute a $p$-dimensional positional encoding, indicating its location in the image, as described in the next section. This is concatenated with the object's visual descriptor. 

We also compute a visual descriptor and positional embedding for the target object. If there are $m$ objects in the image and one target, we then have a set of object embeddings $F_{e} \in \mathbb{R} ^ {(m+1) \times (2p)}$ after the object embedding module.

Subsequently, the set of embeddings $F_{e}$ is fed into $n_e$ stacked identical transformer encoder blocks. The encoder outputs the hidden states of the sequence, $H_{e} \in \mathbb{R}^{(m+1) \times h} $, where $h$ is the dimension of hidden states of encoder blocks.

\paragraph{Distance-based Positional Encoding.}

\begin{figure}[t]
 \includegraphics[width=0.4\textwidth]{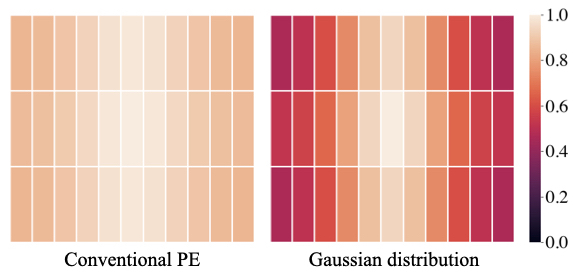}
\centering
\caption{(left) Cosine similarity of the PE at the centre position and the PEs at other positions for the PE in \cite{vaswani2017attention}. (right) A Gaussian distribution.
}
\label{pe}
\end{figure} 

Conventional fixed positional encoding (PE) algorithms~\cite{vaswani2017attention,dosovitskiy2020image} use sinusoidal functions to generate positional encodings. 
While these encodings can differentiate between different locations, they are not designed to encode information about the spatial distance between different locations. 
Figure \ref{pe} (left image) illustrates the cosine similarity between the PE of the centre position and the PE of the other positions in a 2D space for the PE defined in \cite{vaswani2017attention}. While there is some decay in similarity as the distance from the center increases, this decay is slow. Thus, although there are differences between PEs at locations close and far from the center, these are small. 
Spatial positions would be better encoded if cosine similarity between PEs decayed more rapidly, such as by following the Gaussian distribution as illustrated in Figure \ref{pe} (right image).
While learnable positional encodings might be able to capture this information, they are not explicitly designed to do so and thus would require inordinate amounts of training data to learn this if it is relevant to the task. 




Spatial distance is an important factor in determining saccades. Most saccades are to close objects (see Figure \ref{pedis}). Large saccades are relatively rare.
Given the importance of spatial distance in determining scanpath behavior, we propose learnable PE that more explicitly encodes spatial distance between locations. We consider an encoding of three spatial dimensions: $(x,y)$ as image locations, and $z \in \{0,1\}$ as a target indicator, constructed by concatenating three 1D PEs.


For the 1D PE, we learn the set of PEs $P_i \in \mathbb{R} ^ {L \times \frac{p}{3} }$ where $i$ indicates spatial location by ensuring that cosine similarity between PEs at different locations decays approximately following a zero-mean Gaussian distribution, by minimizing 
\begin{equation}    
\mathbf{l_{PE}} = \frac{2}{L \times (L-1)} \sum_{i=0}^{L} \sum_{j=i+1}^{L} (\frac{P_i  P_j}{|P_i| |P_j|} - f(i, j))+\lambda r(P),
\end{equation}
where $f(i, j)=
e^{-\frac{1}{2}\left(\frac{|i-j|}{\sigma}\right)^{2}}$, $r(\cdot)$ is the regularization term that constrains each vector from the learned PE matrix to have a unit magnitude and $\lambda$ is the weighting factor. $\sigma$ denotes the standard deviation of the Gaussian distribution. 

We scale the PEs by $\alpha$, to make their magnitude similar to the magnitude of the visual feature branch in the object embedding module. For an object with 2D coordinates $(x, y)$, its positional embedding is $P(x) \oplus P(y)$, where $\oplus$ stands for the concatenation. Since we have an extra dimension $z$ representing the target, we also concatenate the embedding with $P(z)$.


\paragraph{Decoder.}
These $n_d$ decoder blocks have a similar structure to the encoder blocks, except that the decoder has cross-attention layers, where the keys and values come from the encoder output and queries come from the decoder input. The output of the decoder is the hidden states $H_{d} \in \mathbb{R} ^ {l \times h}$. Then we input the hidden state of the last fixation  $H_d^{last} \in \mathbb{R} ^ {1 \times h}$ and all the hidden states from the encoder $H_e$ into the object attention modelling module.

\paragraph{Object Attention (OA)}

In this final step, we model object attention based on the hidden states of the encoder and decoder. $H_e$ serves as key and $H_d^{last}$ is query. The feed-forward layer computes attention embedding of the encoder and decoder, $A_e \in \mathbb{R} ^ {(m+1) \times k}$  and $A_d \in \mathbb{R} ^ {1 \times k}$, respectively, where $k$ is the attention embedding size. Then it outputs the multiplication of two features: $O = \frac{A_e \times A_d.T}{\sqrt{h}} $. We take the average of $n_c$ outputs from $n_c$ identical OA blocks.

Finally, we apply a Softmax to the output of the last module and obtain a probability distribution $D \in \mathbb{R} ^ {m+1}$, where the extra probability represents the end token. If the end token has the largest probability, the generation is terminated. The next fixation can be either taken as the one with the largest probability (greedy sampling) or taken by sampling randomly from this distribution.

\paragraph{Training and Testing.}
The OAT takes as input a target object and an image containing the target and distractors. It outputs a gaze scanpath prediction.

During training, the input to the decoder is shifted because we append a start token [BOS] to the beginning of the sequence. At each step, we compute the cross-entropy loss between the output distribution and the ground truth distribution at that step (a binary matrix where only one value is one). Due to the nature of our task, we do not apply any pre-training or data augmentation.
For inference, the decoder starts from [BOS] and predicts subsequent outputs iteratively, as it takes the newly generated output as the next input.

\section{Experimental Results}

To assess the performance of OAT, we evaluated it on a public dataset and a dataset we collected. 
To account for the longer and more variable scanpaths in these datasets in comparison to datasets with less clutter, like COCO-Search18, we introduce a new metric based on human behavior. We also report performance using common metrics used previously to facilitate comparison with other approaches. A series of experiments were conducted to validate the interpretability and generalization of the proposed model.

\subsection{Datasets}


\begin{table}[t]
\centering
\setlength{\tabcolsep}{1.2mm}{}
\caption{Dataset Summary}
\scalebox{0.71}{
\begin{tabular}{l|l|l|l|l|l|l|l}
\midrule[1.2pt]
                  &  \begin{tabular}[c]{@{}l@{}}\# Target \\ category\end{tabular} & Shelf size     & \# Images & \# Scanpaths & Mean/Std. length & Target area   & Resolution  \\
                  \midrule[1.2pt]
Amazon book cover & 5                  & 6 * 14         & 100       & 585          & 12.9/9.1         & 1.2\%         & 1600 * 2554 \\
\midrule[0.6pt]
Yogurt/wine       & 40                 & 3 * 9 / 2 * 11 & 196       & 858          & 7.7/3.3          & 3.2\% / 3.9\% & 1050 * 1680 \\
\midrule[1.2pt]
\end{tabular}}
\label{dataset}
\end{table}

\begin{figure*}[t]
\includegraphics[width=0.8\textwidth]{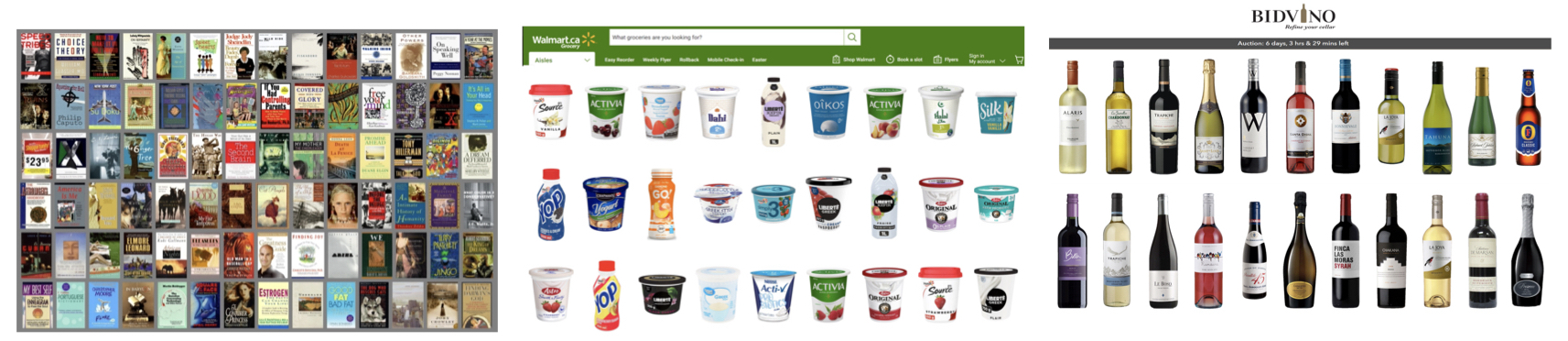}
\centering
\caption{Example product array images.}
\label{data}
\end{figure*} 

We trained and evaluated OAT on two datasets. The summary of the datasets and examples images are shown in Table \ref{dataset} and Figure \ref{data}, respectively. The first is the Amazon book cover dataset \cite{Stauden18}. It contains 102 images of 6 by 14 arrays of randomly shuffled book covers. The dataset provides fixation data from six participants on average per image searching for a target book cover. To study the applicability of our model to other scenarios, we collected another similar dataset, but with 2D arrays of yogurt and wine. The wine product arrays have 2 $\times$ 11 packages from 21 brands. The yogurt product array includes 3 $\times$ 9 packages from 13 brands. Each subject viewed around 30 shelf-target combinations. More details about the dataset collection are provided in the supplementary materials.
            
We convert the fixations into their corresponding objects based on bounding boxes placed around the objects.

\subsection{Implementation Settings}

We used four encoder and decoder blocks and two OA blocks. Each layer contained four attention heads. The $p$ of the patch embedding module is 128. The size of hidden states $h$ was 256. The $\mu$ and $\sigma$ parameters in the PE loss function were 0 and 2, respectively. We set $\lambda$ equal to 1 and $L$ equal to 11, the largest dimension size of the shelf in our dataset. The PE matrix was trained with a learning rate of 0.01 and 10000 iterations. We trained OAT with a batch size of 20 and a learning rate of 1e-4. The data was randomly split into 8:1:1 as training/validation/testing data. 
                          
\subsection{Metrics}
Human scanpaths for the same stimulus vary from trial to trial. Therefore, for all the baselines and OAT, we generated 100 trajectories by sampling from the output distribution and averaged the results across these trajectories. We evaluated the models using the following four metrics.

(i) \textit{Sequence Score (SS) \cite{yang2020predicting}:} SS is commonly used in pixel-level scanpath prediction. It clusters fixations and converts the output sequence of fixations into cluster IDs, and then uses a string matching algorithm to measure the similarity between
two strings. Since we have object-level outputs, we treat object output as the cluster ID and directly use the string-matching algorithm.

(ii) \textit{Fixation Edit Distance (FED) \cite{mondal2023gazeformer}:} Similar to the SS, the FED converts scanpaths to strings but uses the Levenshtein distance \cite{levenshtein1966binary} for string-matching.

(iii) \textit{Behavioural measures:} 
It has been observed that individuals often revisit previously inspected items when searching for a specific target due to the inherent limitations of working memory, necessitating the need for rehearsal \cite{beck2006memory}. Commonly used string-matching algorithms, fail to adequately account for the distinct behavioural patterns associated with the retrieval of new and previously encountered objects. This aspect has been overlooked in previous research and warrants comprehensive evaluation. 

To this end, we propose a new metric to measure these behavioural patterns. Gaze patterns involve three attention states: 
\textbf{search} (viewing a new object for exploration), \textbf{revisits} (re-viewing an object), and successive \textbf{re-fixations} on the same object to confirm its identity \cite{niebur1998computational}.We propose to quantify the extent to which OAT conforms to human behaviour by comparing the percentages of search, revisits, and refixations; the accuracy of finding the correct target (Accuracy); the average length of the scanpaths in fixations (Avg. Length); and the average absolute percentage difference between these values and human scanpaths (Overall). Mathematically, for one predicted scanpath $\mathcal{S} = \{O_1, O_2, ..., O_n\}$ which consists of $n$ objects, we calculate the length $n$, the target accuracy $\mathbbm{1}(O_n = T)$ where $T$ is the target object, search $\frac{a}{n-1}$, revisit $\frac{b}{n-1}$, and refix $\frac{c}{n-1}$. For $n-1$ saccades in the scanpath, $a+b+c=n-1$. $a$ is the number of fixations on objects that were not viewed before. $b$ is the number of fixations on objects that have already been viewed but not in the last fixation, and $c$ is the number of fixations on the same object as the last fixated one.

(iv) \textit{Multi Match (MM) \cite{dewhurst2012depends}:} 
To assess the performance on pixel level with other pixel-level based baselines, we convert the object-level prediction of OAT to the pixel-level predictions, by assuming the fixations fall on the center pixels of the objects, and compare this pixel sequence with the pixel-level outputs of other baselines. MM measures the scanpath similarity from four perspectives: shape, direction, length, and position scores.


\subsection{Quantitative Performance Comparison }

To assess the efficacy of OAT, we conducted a comparative analysis with four baseline methods. We considered two types of baseline methods: 

(i) IID random sampling:
Fixations are randomly selected from one of two distributions over the objects independently for each fixation. Stopping is determined by a geometric random process where the stopping probability is determined by the average sequence length in the training dataset. The two distributions were: (1) \textit{Random} (Uniform over all objects) (2) \textit{Center} (Decaying with distance from the centre following a Gaussian distribution~\cite{judd2009learning}. 

(ii) Temporal modelling of fixation probabilities conditioned on previous fixations: 
   (1) \textit{WTA (Winner-take-all \cite{koch1985shifts})}. A saliency map is computed as in \cite{itti1998model}, and the initial fixation is determined at the location of maximum saliency. Inhibition-of-return is applied and subsequent fixations are selected based on the maximum saliency within the updated map. This process is repeated iteratively until the length reaches the average sequence length in the training dataset.
    (2) \textit{IRL \cite{yang2020predicting}}: IRL applies inverse reinforcement learning to scanpath prediction and learns a target-related reward function. 
    (3) \textit{Gazeformer \cite{mondal2023gazeformer}}: The Gazeformer is a transformer-based architecture. It applies a language model (LM) to generate an embedding of a textual description of the target and fuses it with the global features from the image. Since the targets in our datasets are specified as images, rather than text, we applied image captioning to each target image using Blip \cite{li2022blip} to generate a textual description of each target. 

The baselines output the sequences of pixel-level fixations. We convert pixel-level fixations to object-level fixations according to the object bounding boxes into which the pixel-level fixation fall into, except when calculating the MM metric.

\begin{table*}[t]
\centering
\small 
\setlength{\tabcolsep}{1.3mm}{}
\caption{Comparison with baselines in Amazon book cover dataset. Percentages in parentheses indicate the percentage change relative to human behaviour.}
\scalebox{0.75}{
\begin{tabular}{l|cccccc|ccccc|cc}
\midrule[1.2pt]
              & \multicolumn{6}{c|}{\textbf{Behaviour Difference}}     &  \multirow{2}{*}{\begin{tabular}[c]{@{}c@{}} \\ \textbf{FED} $\downarrow$ \end{tabular}} &  \multirow{2}{*}{\begin{tabular}[c]{@{}c@{}} \\ \textbf{SS} $\uparrow$ \end{tabular} }  \\ 
              & Search (\%) & Revisit (\%) & Refix (\%) & \begin{tabular}[c]{@{}c@{}}Accuracy \\  (\%) \end{tabular} & \begin{tabular}[c]{@{}c@{}}Avg. Length \\ (\# fixations) \end{tabular}  &\textbf{Overall} $\downarrow$                               &                               \\ \midrule[1.0pt]
Human         & \underline{85.8}  & \underline{2.3}   & \underline{11.9}                                               & \underline{91.7} & \underline{8.4}                                             & 0                       & 0        & 1                                                       \\ \midrule[0.6pt]
Random        & 95.5 (+0.11) & 3.6 (+0.56)  & 0.9 (-0.92)  & 1.1 (-0.99)  & 8.8 (+0.05)                                       & 0.530       & 11.73                   & 0.042                                \\
Center \cite{judd2009learning}    &86.5(+0.01)   &  10.3(+3.48)&  3.2(-0.73)  &   1.3 (-0.98)    & 8.9(+0.06)  &   1.053                        &  11.73          &   0.059                                        \\
WTA \cite{koch1985shifts, itti1998model}           &  84.3(-0.02)    &   12.0(+4.22)     &  3.7(-0.69)    & 1.2(-0.99)    &  8.8(+0.05)   & 1.186   &  11.60     & 0.056                                                             \\
IRL \cite{yang2020predicting} & 83.4 (-0.03) & 14.1 (+5.13) & 2.5(-0.79) & 2.5(-0.97) & 12.3(+0.46) & 1.466 & 13.70 & 0.051 \\
Gazeformer \cite{mondal2023gazeformer}    &  81.2 (-0.05) & 3.2(+0.39) & 15.6(+0.31) & 1.7(-0.98) & 6.0(-0.29) &      0.404 & 8.65 & 0.098                         \\ \midrule[1.0pt]
\textbf{OAT}  & \underline{85.3}(-0.01)  & \underline{3.0} (+0.30)    & \underline{11.7} (-0.02)          & \underline{89.4} (-0.03)  & \underline{8.5}(+0.01)   &  \textbf{0.074}        &\textbf{0.68}             & \textbf{0.950}                       \\ \midrule[1.2pt]                
\end{tabular}}
\label{table1}
\end{table*}

\begin{table*}[t]
\centering
\small 
\setlength{\tabcolsep}{1.3mm}{}
\caption{Comparison with baselines in our collected dataset. Percentages in parentheses indicate the percentage change relative to human behaviour.}
\scalebox{0.75}{
\begin{tabular}{l|cccccc|ccccc|cc}
\midrule[1.2pt]
              & \multicolumn{6}{c|}{\textbf{Behaviour Difference}}     &  \multirow{2}{*}{\begin{tabular}[c]{@{}c@{}} \\ \textbf{FED} $\downarrow$ \end{tabular}} &  \multirow{2}{*}{\begin{tabular}[c]{@{}c@{}} \\ \textbf{SS} $\uparrow$ \end{tabular} }  \\ 
              & Search (\%) & Revisit (\%) & Refix (\%) & \begin{tabular}[c]{@{}c@{}}Accuracy \\  (\%) \end{tabular} & \begin{tabular}[c]{@{}c@{}}Avg. Length \\ (\# fixations) \end{tabular}  &\textbf{Overall} $\downarrow$                               &                               \\ \midrule[1.0pt]
Human         & \underline{63.9}  & \underline{4.4}   & \underline{31.7}                                               & \underline{84.9} & \underline{8.1}                                             & 0                       & 0        & 1                                                       \\ \midrule[0.6pt]
Random        & 88.6(+0.39) & 8.6 (+0.95)   & 2.8 (-0.91)                                                & 4.3 (-0.95)  & 7.6 (-0.06)                                         & 0.657          & 10.09                   & 0.093                                  \\
Center \cite{judd2009learning}       & 85.4 (+0.34)  & 10.5 (+1.39)   & 4.0 (-0.87)                                               & 4.3 (-0.95)   & 7.7 (-0.05)                                           &     0.719       &9.98               & 0.119                                        \\
WTA \cite{koch1985shifts, itti1998model}           &  72.0 (+0.13)      &  18.8 (+3.27)      &  9.2 (-0.71)      &    3.3 (-0.96)                                              & 7.7 (-0.06)                                                   & 1.032              & 10.05              &  0.097                                                              \\
IRL \cite{yang2020predicting} & 83.1(+0.3) & 14.4(+2.27) & 2.5(-0.92) & 49.0(-0.42) & 12.4(+0.54) & 0.901 & 11.72 & 0.193 \\
Gazeformer \cite{mondal2023gazeformer}    &  29.9(-0.53) & 6.7(+0.52) & 63.4(1) & 9.3(-0.89) & 7.2(-0.11) &       0.614 & 7.69 & 0.177                         \\ \midrule[1.0pt]
\textbf{OAT} & \underline{62.9} (-0.02)  & \underline{4.0} (-0.09)   & \underline{33.0} (+0.04)             & \underline{85.7} (+0.01) & \underline{7.8} (-0.04)  &  \textbf{0.031}        &\textbf{7.19}             & \textbf{0.346}                       \\ \midrule[1.2pt]                
\end{tabular}}
\label{table2}
\end{table*}

\begin{table}[]
\setlength{\tabcolsep}{3mm}{}
\caption{Multi Match comparison in Amazon book cover/our collected dataset}
\centering
\scalebox{0.75}{
\begin{tabular}{l|lllll}
\midrule[1.2pt]
             & \textbf{Vector} & \textbf{Direction} & \textbf{Length} & \textbf{Position} & \textbf{Average} \\ \hline
Random       & 0.846/0.866           & 0.603/0.593              & 0.786/0.802           & 0.675/0.687             & 0.727/0.737            \\
Center & 0.906/0.882 &  0.621/0.601 & 0.886/0.830 & 0.711/0.708  & 0.781/0.755 \\
WTA         & 
0.910/0.918 & 0.620/0.609 & 0.892/0.888 & 0.694/0.680 & 0.779/0.774
\\
Gazeformer   & 0.937/\underline{0.956}  & 0.739/0.588              & 0.918/0.922           & 0.828/0.793             & 0.855/0.815            \\
IRL          & 0.861/0.881           & 0.641/\underline{0.645}     & 0.809/0.830           & 0.731/0.744             & 0.761/0.775            \\ \midrule[1.2pt] 
\textbf{OAT}  & \underline{0.978}/0.942           & \underline{0.916}/0.603              & \underline{0.972}/\underline{0.922}  & \underline{0.964}/\underline{0.820}    & \textbf{0.958/0.822}   \\ \midrule[1.2pt] 
\end{tabular}}
\label{tableMM}
\vspace{-4mm}
\end{table}

As shown in Table \ref{table1} (Amazon book cover dataset) and Table \ref{table2} (our collected dataset), OAT outperforms all baseline methods across all metrics. Notably, the behaviour difference between scanpaths generated by the OAT and human scanpaths is only 3.12\% in our collected dataset, underscoring its effectiveness in emulating human attention patterns. IID random sampling shows poor performance across all metrics, due to the absence of temporal dependencies. Among the baseline methods that include temporal dependencies, WTA only updates the distribution by zeroing out the neighbour area of the last viewing point. Thus, it does not capture long-range spatiotemporal dependencies. Because the Gazeformer appends the target to each patch token independently, it does not enable the target representation to be influenced by the global context, nor the integration of global context into local patch representations. In addition, because it predicts all the fixations in one go, it does not account for past gaze history when generating each fixation sequence. While this may be effective when scanpaths are short, e.g., as in COCO-Search18, it fails to capture the long-distance temporal dependencies within individual scanpaths. Similarily, IRL leverages global image features and fails to capture the informative features of the objects. In contrast, our OAT outputs every fixation in a sequence, and incorporates memory reinforcement through OA, allowing for accurate sequence modelling of gaze scanpaths.

Table \ref{tableMM} shows the pixel-level MultiMatch comparison in two datasets. Despite the relatively coarse object-level predictions, our method still outperforms other baselines in average scanpath similarity, since it more accurately estimates the objects being fixated, thus leading to closer pixel-level predictions despite object-level quantization. 

\subsection{Scanpath Visualization in Spatial Dimension}

\begin{figure*}[t]
 \includegraphics[width=0.8\textwidth]{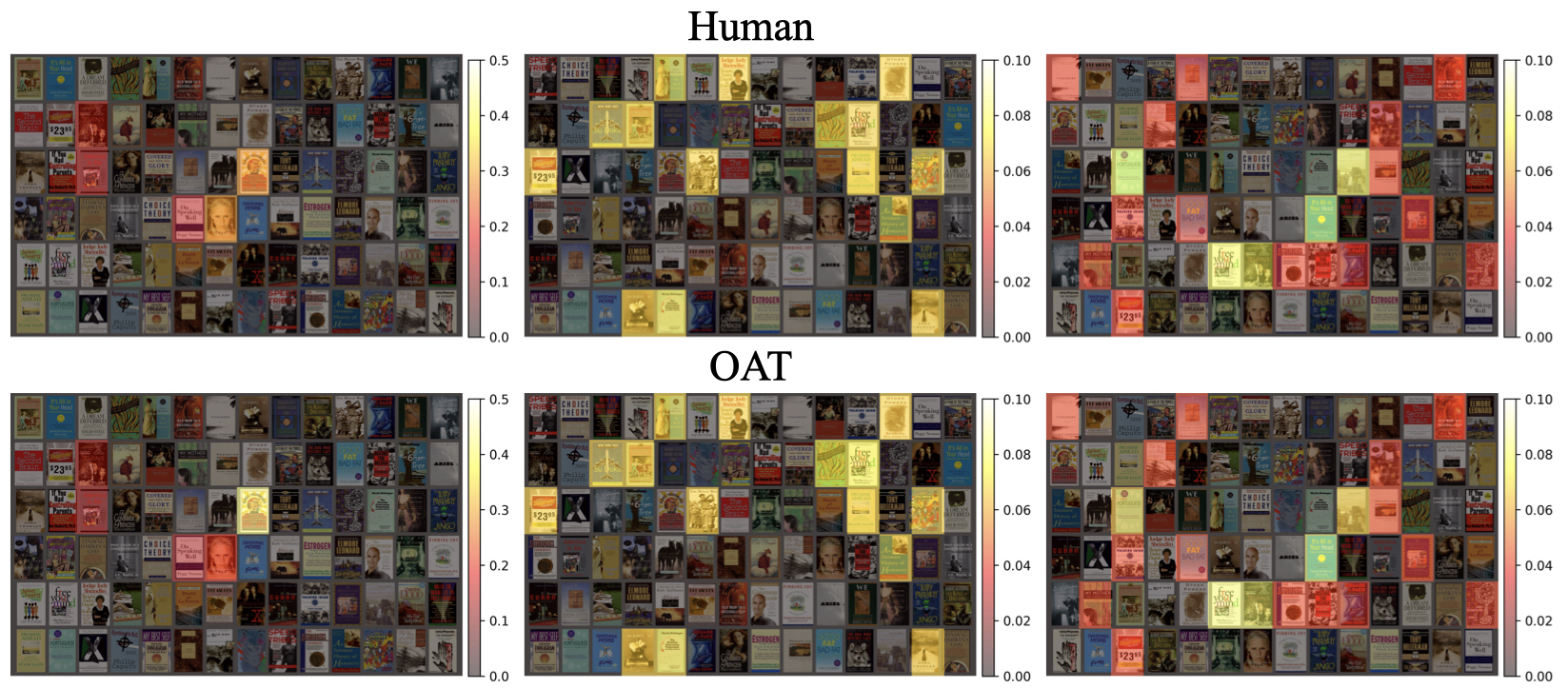}
\centering
 \caption{Heatmap of scanpaths predicted by OAT and generated by human.}
\label{heatmap}
\end{figure*} 

We generated object-level heatmaps for predicted and human scanpaths by calculating the percentage of time each objects was viewed. For OAT, we averaged over 100 predicted scanpaths. The results are visualized in Figure \ref{heatmap}. Predicted heatmaps closely align with the human heatmaps. Only minor discrepancies in the probability distribution are observed between the predicted and human scanpaths. This heatmap analysis shows that OAT can effectively capture object-level attention dynamics.

\subsection{Generalization to Unknown Categories}

\begin{figure*}[t]
 \includegraphics[width=0.6\textwidth]{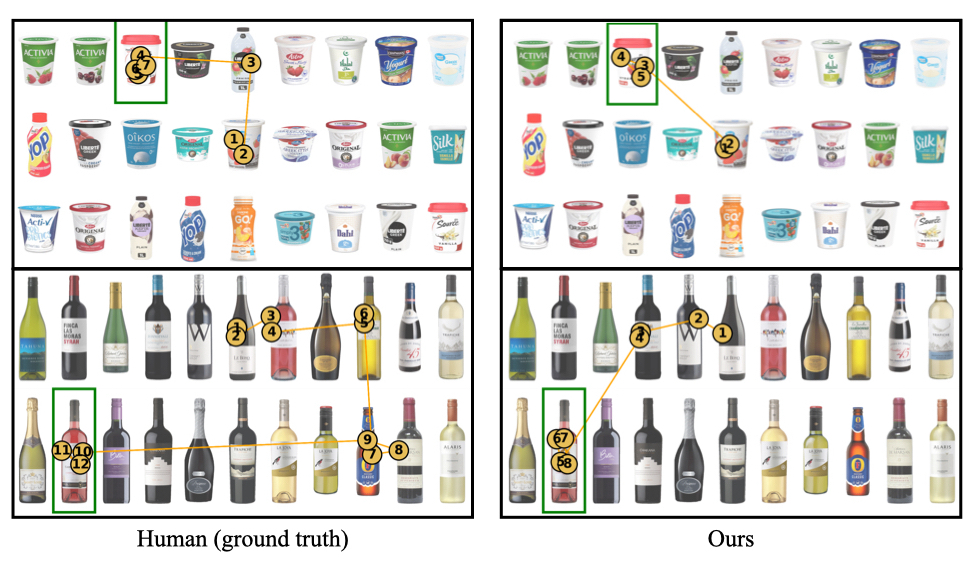}
\centering
 \caption{Generalization ability of OAT.}
\label{comb}
\end{figure*} 

In order to assess OAT's predictions on unseen data, we adopt a random selection approach in our collected dataset whereby a single pair of layout and target object is chosen as the testing dataset, while the remaining dataset is utilized for training. The observed behaviours in Figure \ref{comb} demonstrate that OAT exhibits comparable patterns to those observed in human scanpaths. This outcome not only validates the effectiveness of OAT in accurately modeling consumer scanpaths but also underscores its ability to generalize well to previously unseen layouts and target objects.


\subsection{Parameter Analysis of Distance-based  PE v.s. End-to-End Learnable PE and Relative PE}

\begin{table*}[t]
\centering
\small
\setlength{\tabcolsep}{1.3mm}{}
\caption{Parameter analysis of DPE, E2E and RPE.}
\vspace{-4mm}
\scalebox{0.75}{
\begin{tabular}{l|cccccc|ccccc|cc}
\midrule[1.2pt]
\multicolumn{1}{c|}{}                                                & \multicolumn{6}{c|}{\textbf{Behaviour Difference}}                                                                                                                    & \multirow{2}{*}{\begin{tabular}[c]{@{}c@{}} \\ \textbf{FED} $\downarrow$ \end{tabular}  }  & \multirow{2}{*}{\begin{tabular}[c]{@{}c@{}} \\ \textbf{SS} $\uparrow$ \end{tabular} }
\\ 
\multicolumn{1}{c|}{}                                                & Search (\%) & Revisit (\%)& Refix (\%)  & \begin{tabular}[c]{@{}c@{}} Accuracy \\  (\%)\end{tabular} & \begin{tabular}[c]{@{}c@{}}Avg. Length \\ (\# fixations)\end{tabular} & \textbf{Overall} $\downarrow$  &                              &                               \\ 
\midrule[0.6pt]
Human         & \underline{85.8}  & \underline{2.3}   & \underline{11.9}                                               & \underline{91.7} & \underline{8.4}                                             & 0                       & 0        & 1                                                       \\ \midrule[0.6pt]

E2E PE &  85.2(-0.01) &3.1(+0.35) &11.7(-0.02) & 89.1(-0.03) & 8.7(+0.04) & 0.088  & 2.076 & 0.830 \\

E2E PE + RPE & 85.4(-0.00) &3.0(+0.30) &11.6(-0.03) & 86.9(-0.05) & 8.9(+0.06) & 0.090  & 1.282 & 0.916\\

DPE + RPE & 85.3(-0.01) &3.0(+0.30) &11.7(-0.02) & 81.6(-0.11) & 8.4(0.00) & 0.088  & 2.076 & 0.830\\

\textbf{DPE} & 85.3(-0.01)  & 3.0(+0.30)   & 11.7(-0.02)            & 89.4(-0.03)  & 8.5(+0.01)   &  \textbf{0.074}        & \textbf{0.681}          & \textbf{0.950}                             
\\ \midrule[1.2pt] 
\end{tabular}}
\label{pe_experiment}
\end{table*}

\begin{figure}[t]
      \centering
      \subfloat[Human]{\includegraphics[width=.27\textwidth]{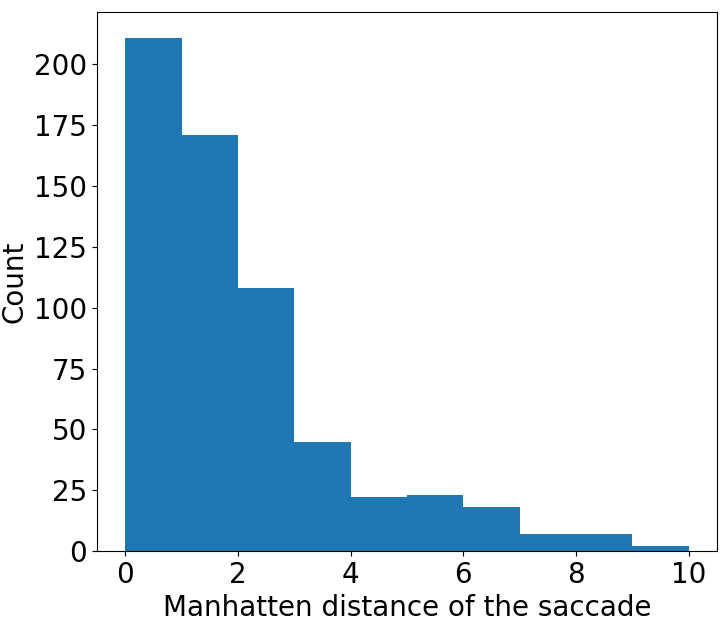}}
      \qquad
      \subfloat[DPE]{\includegraphics[width=.27\textwidth]{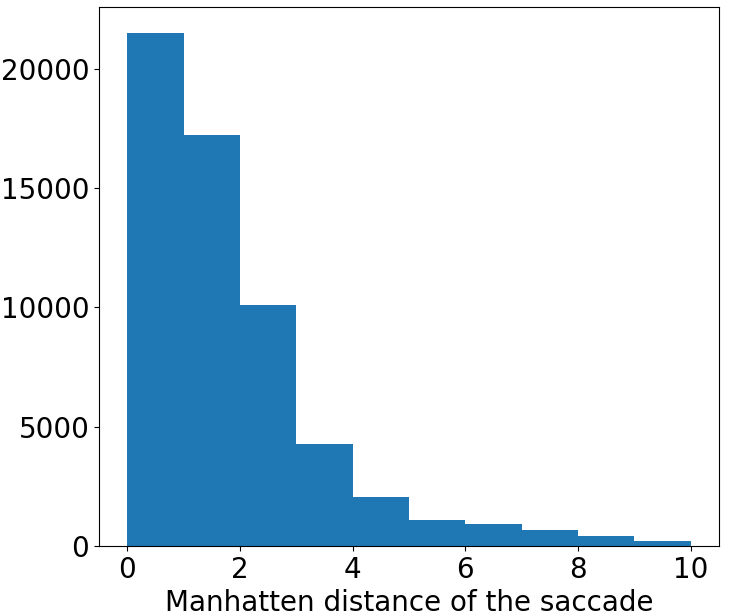}}
      \qquad
      \subfloat[E2E PE]{\includegraphics[width=.27\textwidth]{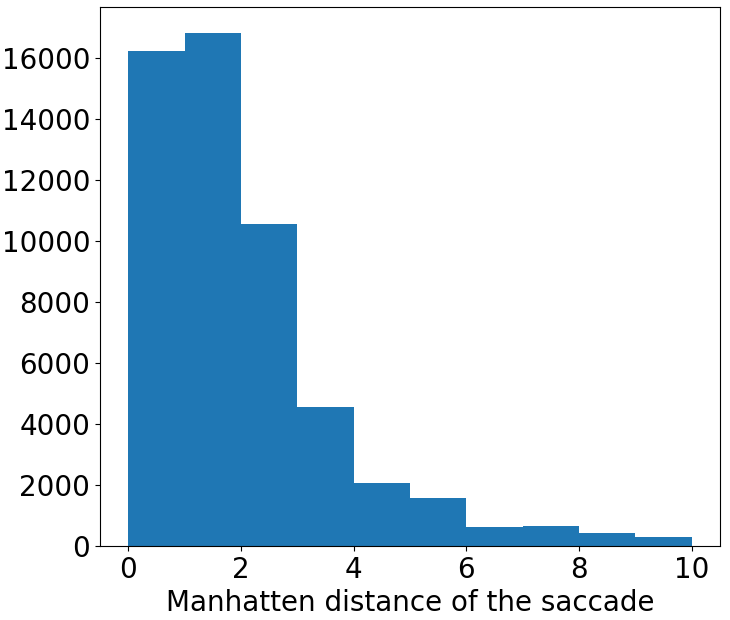}}
  \caption{Distribution of Manhatten distances of the saccades in the predicted scanpaths. Distribution of DPE is closer to human than E2E PE.}
  \label{pedis}
\end{figure}

We compared our proposed Distance-based PE (DPE) with an end-to-end (E2E) learnable PE and Relative PE (RPE) in Amazon book cover dataset to examine the impact of having prior geometric knowledge in the PE component. In the E2E PE approach, we randomly initialized a PE matrix of equivalent size and subsequently trained it in an end-to-end manner as described in \cite{dosovitskiy2020image}. 
RPE \cite{shaw2018self} adds new parameter vectors on top of an existing absolute encoding to encode relative positions. The results in Table \ref{pe_experiment} reveal that DPE outperforms the E2E PE and RPE by 16\% in behavioural metric. 
DPE requires fewer parameters, which gives better performance since gaze training data is limited.
The result suggests that incorporating geometry-based priors helps OAT to mimic the human behaviour of looking at closer objects due to foveated vision. 

To further investigate this effect, we analyzed the distribution of Manhattan distances of all saccades in the predicted scanpaths. In Figure \ref{pedis}, it can be observed that the distribution for the DPE is closer to human behaviour than for the E2E PE. E2E PE generates a larger number of longer saccades, than observed in human data. This suggests that E2E PE does not encode spatial distance as effectively as DPE. Understanding this behavior is crucial for replicating human saccade patterns. Shorter saccades are more prevalent due to energy constraints and the limited information available about peripheral objects. While E2E PE might learn such encoding with sufficient training data, for smaller datasets, incorporating prior knowledge about image geometry via the DPE yields better performance.


\subsection{Importance of object-centric modelling}

To determine whether our model depends upon path tokens encoding single objects, we increased the grid size in the Amazon book covers experiments from 6 $\times$ 14 to 12 $\times$ 28 (so that patches contained only parts of objects) and decreased it to 3 $\times$ 7, so that  (so that patches contained four objects). We mapped fine-scale grid fixations to corresponding objects/groups at coarser scales. The results are shown in Table \ref{ratio}. We see large degradations in performance when patches correspond to parts of objects (Grid 12 $\times$ 28), but relatively small degradations in performance when we model at the object level but consider fixations on groups of objects.

\begin{table}[b]
\centering
\setlength{\tabcolsep}{1.3mm}{}
\caption{Analysis of object-centric attribute (mapped to grid size 6 $\times$ 14 / 3 $\times$ 7)}
\scalebox{0.75}{
\begin{tabular}{l|cccccc|ccccc|cc}
\midrule[1.2pt]
\multicolumn{1}{c|}{}                                            & \multicolumn{6}{c|}{\textbf{Behaviour Difference}}    & \multirow{2}{*}{\begin{tabular}[c]{@{}c@{}} \\ \textbf{FED} $\downarrow$ \end{tabular}  }  & \multirow{2}{*}{\begin{tabular}[c]{@{}c@{}} \\ \textbf{SS} $\uparrow$ \end{tabular} }
\\ 
\multicolumn{1}{c|}{}                                                & Search (\%) & Revisit (\%)& Refix (\%)  & \begin{tabular}[c]{@{}c@{}} Acc \\  (\%)\end{tabular} & \begin{tabular}[c]{@{}c@{}}Avg. Len \\ (\# fixations)\end{tabular} & \textbf{Overall} $\downarrow$  &                              &                               \\ \midrule[1.0pt]
Human         & \underline{85.8/70.0}  & \underline{2.3/5.5}   & \underline{11.9/24.5}                                               & \underline{91.7/93.3} & \underline{8.4/8.4}                                             & 0                       & 0        & 1                                                       \\ \midrule[0.6pt]
\begin{tabular}[c]{@{}l@{}} Grid 12 $\times$ 28 \end{tabular}  &  78.7/64.1 & 11.6/19.8 & 9.6/16.1 & 36.2/39.1 & 15.2/15.2 & 1.14/0.89 & 13.10/12.39 & 0.31/0.35 
\\
\midrule[0.6pt] 
OAT (6 $\times$ 14) & 85.3/73.3  & 3.0/5.5   & 11.7/21.2            & 89.4/91.7  & 8.5/8.2  &  \textbf{0.07}/0.05        & 
\textbf{0.68}/0.81          & \textbf{0.95}/0.90   
\\
\midrule[0.6pt] 
Grid 3 $\times$ 7  & -/69.9 & -/5.5 & -/24.6 & -/92.9 & -/8.5 & -/\textbf{0.01} & -/\textbf{0.78} & -/\textbf{0.93}
\\
 \midrule[1.2pt] 
\end{tabular}}
\label{ratio}
\end{table}

\subsection{Improving revisit/refix behaviour by the cross attention} We found that the cross-attention mechanism is critical in modelling revisits and refixations. For 20 shelf-target pairs and over 100 generated trajectories each, we looked at how the probabilities of fixating on the object fixated in fixation 2 at subsequent fixations (3, 4, 5 and 6) changed when the object in the fixation record at fixation 2 was replaced by a random object. the changes were $-3.42\%$, $18.14\%$, $21.53\%$, and $33.22\%$, respectively. 
The percentage change in probability for fixation 3 was negative ($-3.42\%$), which is consistent with the increased percentage of re-fixations by OAT in Table \ref{table1} and \ref{table2} compared to most baselines except Gazeformer. It was positive for fixations 4 through 6, which is consistent with the decreased percentage of re-visits by OAT compared to all baselines. This brought OAT's behaviour closer to human behaviour. 

\subsection{Ablation Study}

\begin{table*}[t]
\centering
\small
\setlength{\tabcolsep}{1.3mm}{}
\caption{Ablation Study of two novel components. Percentages in parentheses indicate the percentage change relative to human behaviour.}
\scalebox{0.75}{
\begin{tabular}{l|cccccc|ccccc|cc}
\midrule[1.2pt]
\multicolumn{1}{c|}{}                                                & \multicolumn{6}{c|}{\textbf{Behaviour Difference}}                                                                                                                    & \multirow{2}{*}{\begin{tabular}[c]{@{}c@{}} \\ \textbf{FED} $\downarrow$ \end{tabular}  }  & \multirow{2}{*}{\begin{tabular}[c]{@{}c@{}} \\ \textbf{SS} $\uparrow$ \end{tabular} }
\\ 
\multicolumn{1}{c|}{}                                                & Search (\%) & Revisit (\%)& Refix (\%)  & \begin{tabular}[c]{@{}c@{}} Accuracy \\  (\%)\end{tabular} & \begin{tabular}[c]{@{}c@{}}Avg. Length \\ (\# fixations)\end{tabular} & \textbf{Overall} $\downarrow$  &                              &                               \\ \midrule[1.0pt]
\begin{tabular}[c]{@{}l@{}} Human \end{tabular}        & \underline{63.9}  & \underline{4.4}   & \underline{31.7}                                               & \underline{84.9} & \underline{8.1}                                             & 0                  & 0             & 1                                                                                 \\
\midrule[0.6pt]
\begin{tabular}[c]{@{}l@{}} w/o Both \end{tabular}        & 65.5 (+0.03)   &24.0 (+4.45)   & 10.4 (-0.67)    & 41.0 (-0.52)          &  9.9 (+0.22)                               & 0.533      & 11.66       & 0.224                                        \\

\begin{tabular}[c]{@{}l@{}} w/o DPE\end{tabular}        & 59.2 (-0.07)  & 5.5(+0.25)   & 35.3(+0.11)                                & 86.9(+0.02)        & 9.1(+0.12)                                                     &   0.119         & 7.70        & 0.327                                             \\
\begin{tabular}[c]{@{}l@{}} w/o OA\end{tabular} & 61.1 (-4.4) & 8.2(+0.86)   & 32.5(+0.03)                                              & 74.9(-0.12)      & 8.5(+0.05)                                       &   0.169                      &  7.58        & 0.316                                         \\ 
\midrule[1.0pt]

\textbf{OAT} & \underline{62.9}(-0.02)  & \underline{4.0}(-0.09)   & \underline{33.0}(+0.04)            & \underline{85.7}(+0.01)  & \underline{7.8}(-0.04)   &  \textbf{0.031}        & \textbf{7.19}          & \textbf{0.346}                             
\\ \midrule[1.2pt] 
\end{tabular}}

\vspace{-4mm}
\label{ablation}
\end{table*}

We evaluated the impact of our novel components, DPE and OA, through ablation experiments on our dataset in Table \ref{ablation}. Because models without the OA cannot handle different layouts, we trained two separate models for wine and yogurt and calculate the average outcomes. Removing DPE and OA significantly degraded performance. The behavioral difference was 17 times worse. Removing the DPE component alone led to 3.84 times degradation. Removal of OA resulted in a 5.45 times degradation. The DPE component improves the accuracy of the modeled shelf geometry, while the OA component captures dependencies and enhances model flexibility. Combining these components improves overall performance.



\section{Conclusion}
In this paper, we propose a transformer-based architecture OAT for modelling the gaze scanpaths at the object level.
OAT outperforms all prior baselines in modelling human behaviour. 
It also demonstrates a remarkable ability to generalize on unseen layouts and targets. Moving forward, potential extensions to this work could involve applying to searching for objects with non-uniform shapes, which is more common in real world search tasks.



%
%
\bibliographystyle{splncs04}
\bibliography{main}

\end{document}


\title{Supplementary Material} 

\titlerunning{OAT}

\author{Yini Fang\inst{*1}\orcidlink{0009-0008-9478-5545} \and
Jingling Yu\inst{*1}\orcidlink{0000-0002-3076-1287} \and
Haozheng Zhang\inst{2}\orcidlink{0000-0003-1312-4566} \and
Ralf van der Lans\inst{1}\orcidlink{0000-0002-7726-8238} \and
Bertram Shi \inst{1}\orcidlink{0000-0001-9167-7495}
}

\authorrunning{Y. Fang et al.}

\institute{Hong Kong University of Science and Technology, Clear Water Bay, HK  \and
University of Durham, Durham, UK
\\
\email{\{yfangba, jyubj\}@connect.ust.hk}, \email{haozheng.zhang@durham.ac.uk}, \email{rlans@ust.hk}, \email{eebert@ust.hk}
}

\maketitle

We provide more details about our collected dataset in this document.

\section{Preparation}
\begin{figure*}[t]
\includegraphics[width=\textwidth]{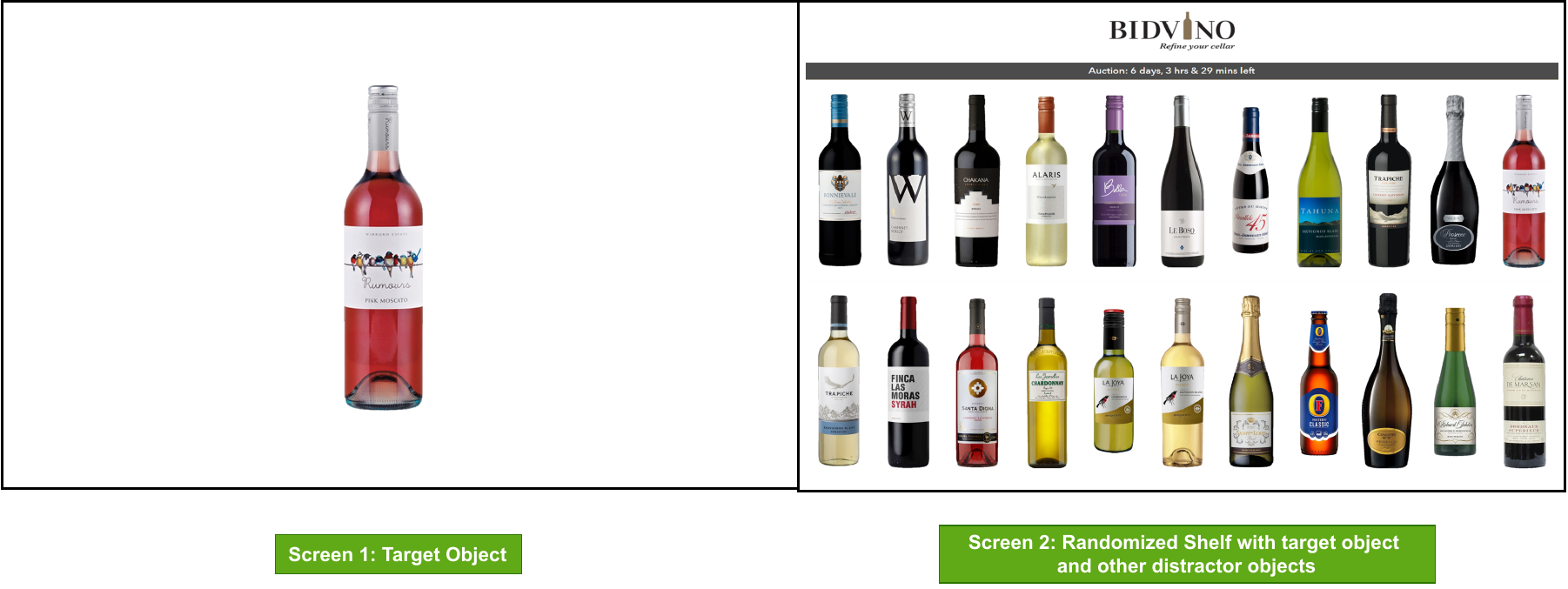}
\centering
\caption{Example of wine stimuli.}
\label{wine}
\end{figure*} 

\begin{figure*}[t]
\includegraphics[width=\textwidth]{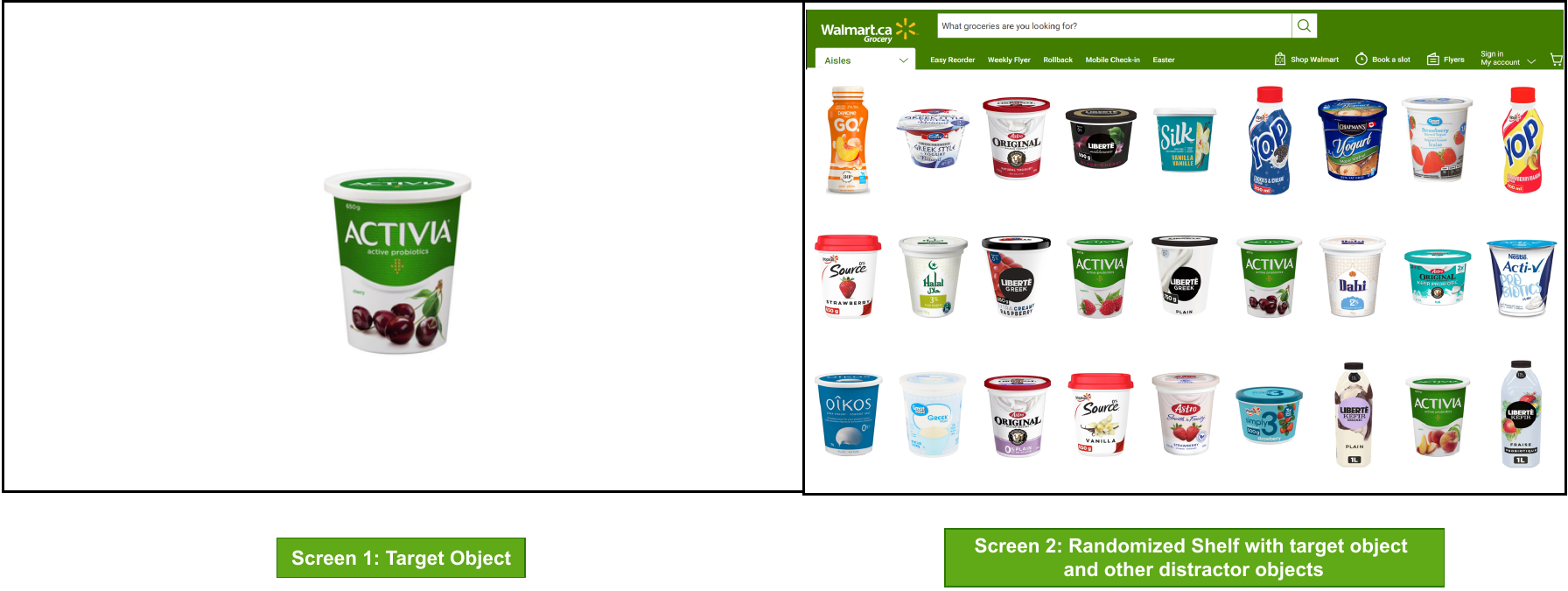}
\centering
\caption{Example of yogurt stimuli.}
\label{yogurt}
\end{figure*} 


Prior to conducting the experiment, a selection of 22 wine products from 21 distinct brands was made from the Bivino website\footnote{https://bidvino.com/}, along with 27 yogurt products from 13 different brands sourced from the Walmart website\footnote{https://www.walmart.com/}. For each type of product, 20 items were designated as target packages, and a total of 100 shelf images were created by arranging these packages in permutations on a two-dimensional grid shelf. The arrangement for the wine products was made into 2 rows and 11 columns, whereas the yogurt products were organized into 3 rows and 9 columns. The images were generated with a resolution of 1680 $\times$ 1050 pixels.

The data collection involved 39 subjects aged between 18 and 30 years old, with 47\% male and 53\% female participants. Eye-tracking data was collected using the Tobii T60XL eye tracker, with a sampling rate of 60Hz.


\section{Collection Process}
\begin{figure*}[t]
\includegraphics[width=\textwidth]{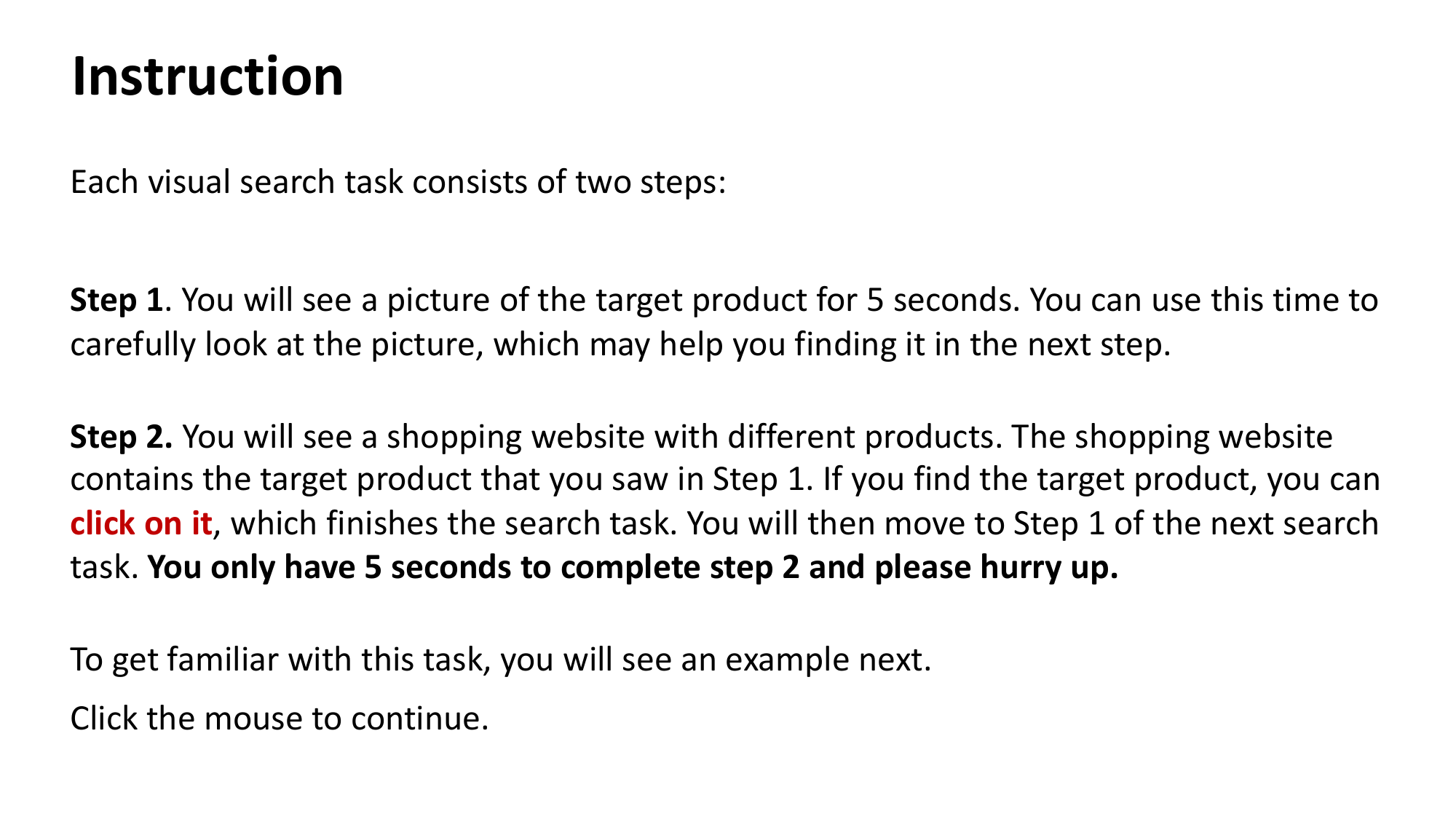}
\centering
\caption{Visual search study instruction.}
\label{instruction}
\end{figure*}

The experimental procedure involved instructing the subjects to search for a specific target product within the given shelf image. A screenshot of the instruction provided to the subjects is shown in Figure \ref{instruction}.

To ensure variability, we randomly assigned 30 unique combinations of target products and shelf images to each subject. In each trial, the target product was initially displayed for a duration of five seconds, followed by the presentation of the corresponding shelf image. The participants were instructed to locate the target product and click on it within a time limit of three seconds. Their compensation was contingent upon both the speed and accuracy of finding the target product.

A total of 1170 scanpaths were collected during the experiment. We utilized the Binocular-Individual Threshold algorithm \cite{van2011defining} to cluster the raw gaze samples into fixations. This algorithm takes into account binocular viewing and leverages the covariation between the two eyes. Subsequently, we performed data cleaning procedures, which involved removing 312 scanpaths with less than 80\% record integrity \cite{van2017eye}. Additionally, redundant fixations were eliminated from the scanpaths after subjects clicked the mouse. Following these postprocessing steps, our dataset consists of 858 scanpaths with an average length of 7.7 fixations.


\section{Consent From Subjects}

Prior to conducting the study, ethical approval was obtained from an Institutional Review Board (IRB) to ensure the appropriateness and adherence to ethical guidelines. Informed consent was obtained from all subjects, who were required to sign a consent form that explicitly outlined the following provisions:

\begin{itemize}
    \item The collected data would be used solely for research purposes.
    \item All participant responses would remain confidential, and no personally identifiable information (such as names or email addresses) would be collected or stored in a manner that could link the responses or eye-tracking data to individual identities.
    \item Only aggregated data would be utilized in any publication that incorporates this dataset.
\end{itemize}

The consent form is included in the appendix. Names and institutions were anonymized for the initial paper review.


%
%
\bibliographystyle{splncs04}
\bibliography{main}